\pdfoutput=1

\documentclass[11pt]{article}

\usepackage{EMNLP2023}

\usepackage{times}
\usepackage{latexsym}

\usepackage[T1]{fontenc}

\usepackage[utf8]{inputenc}

\usepackage{microtype}

\usepackage{inconsolata}

\usepackage{amsmath}
\usepackage{todonotes}

\usepackage{booktabs}
\usepackage{makecell}

\title{From cart to truck: meaning shift through words in English in the last two centuries}

\author{Esteban Rodríguez-Betancourt \and Edgar Casasola-Murillo \\
         Programa de Posgrado en Computación e Informática\\
         Universidad de Costa Rica \\
         San José, Costa Rica \\
         \texttt{\{esteban.rodriguezbetancourt,edgar.casasola\}@ucr.ac.cr}
}

\begin{document}
\maketitle
\begin{abstract}

This onomasiological study uses diachronic word embeddings to explore how different words represented the same concepts over time, using historical word data from 1800 to 2000. We identify shifts in energy, transport, entertainment, and computing domains, revealing connections between language and societal changes.

Our approach consisted in using diachronic word embeddings trained using word2vec with skipgram and aligning them using orthogonal Procrustes. We discuss possible difficulties linked to the relationships the method identifies. Moreover, we look at the ethical aspects of interpreting results, highlighting the need for expert insights to understand the method's significance.

\end{abstract}

\section{Introduction}

Words and their meanings can undergo shifts over time --- a phenomenon referred to as semantic shift. The most common way to study semantic shift is through a \emph{semasiological} perspective \citep{tahmasebi2019survey}: what is studied is how the meaning of a word changes through time. In this paper, we will take a less common route, known as \emph{onomasiological} perspective: we will study how the same concept is represented by different words across time.

In this study, we will use diachronic word embeddings trained with English from 1800 to 2000 to find interesting shifts in the representation of the same concept through time. This approach can complement the more common \emph{semasiological} approaches, and give a better image of society from our current knowledge. For example, we found that in the 1800s, cart had the most similar embedding to the modern embedding of truck.

The article structure is as follows: In the ``Definitions and Previous Work'' section, we explain key terms and touch on previous research. Following this, we outline our methodology. Next, our findings will be presented, emphasizing specific words within the results section. Subsequently, we provide a summarized account of our findings and challenges in the ``Conclusion'' section. Afterward, we address the limitations of our work and engage in an ethics discussion related to this type of study. Additionally, a summary of result tables is available in Appendix \ref{sec:appendixWordAnalogues}.

\section{Definitions and Previous Work}

\emph{Onomasiology} consists in finding which words can be used to represent a given concept or idea. It is the opposite of \emph{semasiology}, which is the study of the meaning of a given word. For example, a typical dictionary can be used to answer \emph{semasiological} questions, as it maps a word with its meaning. \emph{Onomasiology} would be the inverse operation.

The \emph{onomasiology} field started in the late 19th century. Before that, most linguists were interested mostly in the etymology of the words \citep{Grzega2007EnglishAG}. The field not only studied the diachronic shifts of words across time, but also how the same concept was named in different regions.

From the point of view of computational linguistics, it is far easier to find material about \emph{semasiological} shifts than \emph{onomasiological} changes. For example, most of the articles cited in surveys like \citet{kutuzov-etal-2018-diachronic}, \citet{tahmasebi2019survey} and \citet{montanelli2023survey} are focused on how the meaning of a given word changed.

One example of onomasiological study using computational linguistics is the one presented by \citet{szymanski-2017-temporal}. The article introduced the concept of temporal word analogies and was able to identify temporal analogies like ``Ronald Reagan in 1987 is like Bill Clinton in 1997''. He used data from a corpus of New York Times articles, from 1987 to 2007.

Another example is \citet{kutuzov-etal-2019-one}, where the word analogy task was extended to a one-to-X formulation, that allows mapping a relation to nothing. This task was applied to historical armed conflicts and was even used to predict new relations that mapped a location with an armed group.

Similar to the study conducted by \citet{szymanski-2017-temporal}, this paper also identifies temporal analogies; however, it extends the scope over a more extensive historical period. Additionally, rather than focusing on validating the technique against a gold standard, our approach involves leveraging it to uncover words within the ``diachronic neighborhood'' of various meanings of interest, particularly those pertaining to computing, entertainment, transport, and energy. This methodology led us to intriguing findings, such as the relation between the word ``ship'' from the 1800s and the concept of ``aircraft'' in the 1990s, with the two having the most similar embeddings.

Overall, we believe that in the field of language shift research, these types of \emph{onomasiological} studies can complement the more prevalent \emph{semasiological} approach. They provide researchers with a more in-depth insight into the viewpoints of people from the past, potentially facilitating a more accessible understanding of how concepts have evolved by linking historical terminology with modern analogues.

\section{Methodology}

This study aimed to identify instances of word shifts associated with the same concept. To achieve this, the first step was to establish a method for concept representation. Word embedding techniques, such as word2vec \citep{mikolov_efficient_2013}, were employed to map words to dense vectors. These techniques are recognized for their ability to generate embeddings in which words with similar meanings exhibit similar vector representations \citep{le-mikolov-distributed-representations-of-sentences-2014}. Given this capability, we chose to treat word embeddings as representations of the meanings or concepts that we wanted to trace across time.

To conduct the diachronic study, we needed word embeddings trained on data from various periods. For this purpose, we employed the word2vec skip-grams model created by \citep{hamilton-etal-2016-diachronic}, specifically ``All English (1800s-1900s)''. This model comprises English word embeddings for each decade, spanning from the 1800s to the 1990s. The embeddings were trained on Google N-Grams corpus.

\begin{figure}[htbp]
    \centering
    \includegraphics{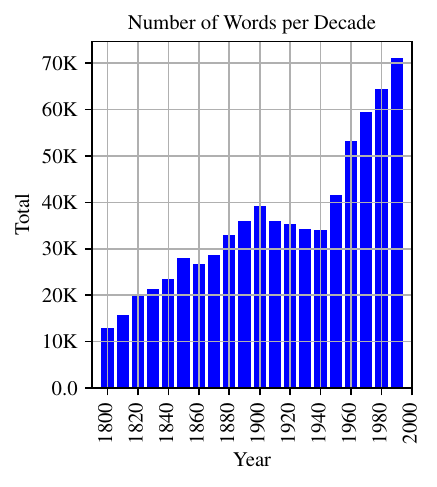}
    \caption{\label{fig:WordsPerDecade}Number of words per decade after cleaning zeroed embeddings.}
\end{figure}

After obtaining the word embeddings, the next step involves their cleaning and alignment. For cleaning, the words associated with an only zero embedding were removed. The final number of words per decade is shown in Figure \ref{fig:WordsPerDecade}. Alignment is the process of determining a matrix rotation that minimizes the distance between corresponding word pairs from different periods. For our study, the alignment was performed using the Orthogonal Procrustes method \citep{Schonemann:1966:GSO}. Specifically, we chose to align the embeddings from each decade with those from the 1990s.

Finally, several concepts were chosen for study. For each concept, we identified the top N words most similar to an embedding at each period. To achieve this, we utilized the \verb|similar_by_vector| function provided by Gensim \citep{rehurek2011gensim}.

\section{Results}
In this section, we will showcase our findings for a selection of words in various domains, including energy, entertainment, computing, and transport. The tables containing summarized results can be found in the appendix (see Appendix \ref{sec:appendixWordAnalogues}).

\subsection{Energy concepts}

The initial set of concepts under study is related to energy: \emph{petroleum}, \emph{diesel}, \emph{electricity}, and \emph{nuclear}. From Table \ref{tab:energy_onomasy}, it is evident that in the 19th century, \emph{coal} and \emph{steam} emerged as the prominent analogues for \emph{petroleum} and \emph{diesel}. This alignment can be attributed to the fact that the industrial revolution heavily relied on steam power, predominantly generated by burning coal. Additionally, related terms such as \emph{boilers} and \emph{engines} are associated with the concept of steam. Notably, the concept of \emph{electricity} displays minimal variation across its analogues, with terms like \emph{electricity}, \emph{electrical}, and \emph{electric} dominating the list. Interestingly, even amidst these electric-related terms, traces of \emph{coal} and \emph{steam} are observable. On the other hand, the term \emph{nuclear} stands out as distinct, primarily linked to war-related vocabulary such as \emph{blockade}, \emph{war}, \emph{alliances}, \emph{piratical}, \emph{explosion}, \emph{projectiles}, \emph{arsenal}, and \emph{weapons}. This suggests that, by the end of the 20th century, the perception of nuclear energy had largely gravitated toward its application in weaponry rather than as a source of energy.

\subsection{Transport concepts}

Under transport, we studied the concepts behind the words \emph{bus}, \emph{truck}, \emph{train} and \emph{aircraft}. The full examples are available in Table \ref{tab:transport_onomasy}. For instance, the modern concept \emph{bus} and \emph{truck} in the early 19th century were associated with transport mediums that used horses or other animals for propulsion means, like \emph{carriages}, \emph{cart} and \emph{wagon}. \emph{Bus} in a single decade was associated with \emph{train}, but it has been associated with \emph{car} since the 1850s, although it clearly were not the same cars as we know currently. In the case of \emph{train}, we found that is it has been associated with the same word since the 1860s, but before was associated with \emph{caravan}, \emph{passengers} and \emph{ride}, among others. The Oxford English Dictionary includes various obsolete definitions of \emph{caravan} that refer to it being a kind of wagon; this likely explains the relation. The concept of \emph{aircraft} had the most changes, as modern airplanes were invented early in the 20th century. Before being associated with the word \emph{aircraft}, the concept was nearer \emph{ships}, \emph{vessel} and \emph{privateers}, suggesting that the modern role of airfare previously was satisfied mostly by seafaring.

\subsection{Entertainment concepts}
Under the category of entertainment, we explored the concepts associated with \emph{radio}, \emph{cinema}, and \emph{television}. Detailed examples are provided in Table \ref{tab:entertainment_onomasy}. Notably, during the early 19th century, terms like \emph{theatres} (with British spelling) and \emph{operas} were found to be analogous to \emph{cinema} and \emph{television}. Analyzing the temporal analogies revealed a shift towards the American spelling ``theater'' after the 1890s, but it could be explained by a higher volume of American text in the training corpus. By the 1920s, \emph{television} became associated with \emph{radio}, and in the 1930s, with \emph{movie}. \emph{Newspapers}, being precursors to these communication mediums, naturally appeared as analogues for both \emph{radio} and \emph{television}. Additionally, in the latter part of the 19th century, \emph{radio} was linked to \emph{telegraph} and \emph{telephone}, both technologies that had gained traction before the onset of radio broadcasting.

\subsection{Computing concepts}

Regarding computing concepts, we chose to study the concepts behind the terms \emph{computer}, \emph{internet}, and \emph{email}. The concept behind \emph{computer} in the 19th century was associated with mathematics and science terms, such as \emph{mathematical}, \emph{science}, \emph{experimental}, \emph{telegraphy}, \emph{chemical}, \emph{engine}, and \emph{laboratory}. It became associated with \emph{computer} during the 1940s. In the case of \emph{internet}, we can observe its association with \emph{information} and \emph{access}, followed by \emph{telegraph}, \emph{mail}, \emph{telephone}, and \emph{television}. Finally, for \emph{email}, it was linked to words related to its physical counterpart, such as \emph{gazette}, \emph{messages}, and \emph{courier}. Subsequently, it became associated with \emph{telegram} and \emph{phone}. Interestingly, the concept of \emph{email} was juxtaposed with \emph{penguin} in the 1960s for some reason. By the end of the 20th century, the \emph{email} concept was aligned with \emph{telex} and \emph{fax}.

\section{Conclusion}
In this study, we've explored onomasiology's potential to reveal shifts in word meanings over time. Through temporal analogies, we've investigated energy, transport, entertainment, and computing domains, uncovering connections between concepts across different eras.

Our analysis highlights language adaptation to technology, society, and perception changes. For example, we can see how the society evolved from using coal and steam to using hydrocarbons or how the theater, telegraph, and cinema concepts were interlinked.

However, cautious interpretation is crucial. Our method enriches historical understanding, yet historical biases are present. Expert insights from areas like linguistics, history, and sociology are required to interpret the results correctly in context. Regardless of that, we confirm that this technique can be used to get useful insights from our society in the past, so its usefulness is not restricted to linguistics.

\section*{Limitations}

This study was conducted using a single dataset and focused exclusively on one language. The utilized word embeddings were trained on the Google Books N-Gram corpus, which is known to possess a bias towards scientific literature \citep{hengchen_challenges_2021}. Therefore, it's important to recognize that this dataset may not offer a fully representative or randomly sampled reflection of the entire English language. Furthermore, the analogies presented in this article might not universally apply to speakers of other languages.

Moreover, the embeddings used were trained in discrete ten-year periods. Consequently, the granularity of these timeframes might not adequately capture nuanced shifts in meaning. A potential approach to addressing this limitation is to repeat the study with finer time intervals, which could unveil more subtle changes.

While the decision to employ word2vec skip-gram embeddings was motivated by their availability and ease of training, it's worth noting that more advanced contextual embedding models like BERT \citep{devlin-etal-2019-bert} are now accessible. However, the utilization of these newer models comes with a trade-off between their enhanced capabilities and the available resources.

\section*{Ethics Statement}

The methods used in this study can help people to get a better grasp of the perspectives of people in the past. However, it may be prone to misunderstandings or may require further analysis by experts to accurately interpret the relations revealed through this method. In this article, we opted to show examples primarily related to technological advancements, which are unlikely to be deemed offensive. We also applied the method to words related to political and social movements, and found that some analogies may be challenging to explain. These challenges could stem from biases held by the authors of the texts used to train the embeddings model, or they might even be inaccurate. We will address these situations in the following paragraphs.

As mentioned, some analogues found can be hard to explain and may require expert knowledge in history, sociology, anthropology or other areas of study to explain what that analogue makes sense, given the context of the people who wrote the texts that were used to build the word embeddings model. For example, it is relatively straightforward to understand the analogy between \emph{steam} in the 1800s and \emph{diesel} in the 1990s, given that the industrial revolution was powered by steam. However, some other analogies we discovered were more challenging to explain, despite the potential historical reasons for their association. For example, \emph{nationalism} ends up being an analogue to several religious groups. From history, we know that in some cases some religious groups contributed to the development of national identities, so the relation may make sense. But it can also be an issue in the model or a bias captured by it.

Another potential concern is that the model may inadvertently capture the prejudices and biases of the authors of the training texts. For example, it is well-documented that word embedding models can inadvertently incorporate gender biases \citep{Bolukbasi2016man}. While allowing models to capture these biases could aid in understanding historical perspectives, it remains the responsibility of researchers to account for these biases when interpreting the generated data and adjust their conclusions accordingly.

Lastly, there is the possibility that the model may generate false analogies due to defects in the training or its inputs. Hence, it is important not to simply accept the outputs of such models, but rather to apply expert criteria to correlate their findings with existing curated knowledge.

Although the employed method is valuable, the analogies generated by the model demand careful evaluation and should not be taken at face value. For instance, what if the model suggests an analogy between a specific population and a contemporary political regime that is currently unpopular? This situation carries the risk that such findings might be misused to justify hostility or discrimination against that particular population. In conclusion, it is crucial to approach these findings with caution and emphasize responsible interpretation to prevent any potential misuse.



\begin{thebibliography}{14}
\expandafter\ifx\csname natexlab\endcsname\relax\def\natexlab#1{#1}\fi

\bibitem[{Bolukbasi et~al.(2016)Bolukbasi, Chang, Zou, Saligrama, and
  Kalai}]{Bolukbasi2016man}
Tolga Bolukbasi, Kai-Wei Chang, James Zou, Venkatesh Saligrama, and Adam Kalai.
  2016.
\newblock Man is to computer programmer as woman is to homemaker? debiasing
  word embeddings.
\newblock In \emph{Proceedings of the 30th International Conference on Neural
  Information Processing Systems}, NIPS'16, page 4356–4364, Red Hook, NY,
  USA. Curran Associates Inc.

\bibitem[{Devlin et~al.(2019)Devlin, Chang, Lee, and
  Toutanova}]{devlin-etal-2019-bert}
Jacob Devlin, Ming-Wei Chang, Kenton Lee, and Kristina Toutanova. 2019.
\newblock \href {https://doi.org/10.18653/v1/N19-1423} {{BERT}: Pre-training of
  deep bidirectional transformers for language understanding}.
\newblock In \emph{Proceedings of the 2019 Conference of the North {A}merican
  Chapter of the Association for Computational Linguistics: Human Language
  Technologies, Volume 1 (Long and Short Papers)}, pages 4171--4186,
  Minneapolis, Minnesota. Association for Computational Linguistics.

\bibitem[{Grzega and Sch{\"o}ner(2007)}]{Grzega2007EnglishAG}
Joachim Grzega and Marion Sch{\"o}ner. 2007.
\newblock \href {https://www1.ku.de/SLF/EngluVglSW/OnOnMon1.pdf} {\emph{English
  and General Historical Lexicology : Materials for Onomasiology Seminars}}.
\newblock Katholische Universität Eichstätt.

\bibitem[{Hamilton et~al.(2016)Hamilton, Leskovec, and
  Jurafsky}]{hamilton-etal-2016-diachronic}
William~L. Hamilton, Jure Leskovec, and Dan Jurafsky. 2016.
\newblock \href {https://doi.org/10.18653/v1/P16-1141} {Diachronic word
  embeddings reveal statistical laws of semantic change}.
\newblock In \emph{Proceedings of the 54th Annual Meeting of the Association
  for Computational Linguistics (Volume 1: Long Papers)}, pages 1489--1501,
  Berlin, Germany. Association for Computational Linguistics.

\bibitem[{Hengchen et~al.(2021)Hengchen, Tahmasebi, Schlechtweg, and
  Dubossarsky}]{hengchen_challenges_2021}
Simon Hengchen, Nina Tahmasebi, Dominik Schlechtweg, and Haim Dubossarsky.
  2021.
\newblock \href {https://doi.org/10.48550/arXiv.2101.07668} {Challenges for
  {Computational} {Lexical} {Semantic} {Change}}.

\bibitem[{Kutuzov et~al.(2018)Kutuzov, {\O}vrelid, Szymanski, and
  Velldal}]{kutuzov-etal-2018-diachronic}
Andrey Kutuzov, Lilja {\O}vrelid, Terrence Szymanski, and Erik Velldal. 2018.
\newblock \href {https://aclanthology.org/C18-1117} {Diachronic word embeddings
  and semantic shifts: a survey}.
\newblock In \emph{Proceedings of the 27th International Conference on
  Computational Linguistics}, pages 1384--1397, Santa Fe, New Mexico, USA.
  Association for Computational Linguistics.

\bibitem[{Kutuzov et~al.(2019)Kutuzov, Velldal, and
  {\O}vrelid}]{kutuzov-etal-2019-one}
Andrey Kutuzov, Erik Velldal, and Lilja {\O}vrelid. 2019.
\newblock \href {https://doi.org/10.18653/v1/W19-4724} {One-to-{X} analogical
  reasoning on word embeddings: a case for diachronic armed conflict prediction
  from news texts}.
\newblock In \emph{Proceedings of the 1st International Workshop on
  Computational Approaches to Historical Language Change}, pages 196--201,
  Florence, Italy. Association for Computational Linguistics.

\bibitem[{Le and
  Mikolov(2014)}]{le-mikolov-distributed-representations-of-sentences-2014}
Quoc Le and Tomas Mikolov. 2014.
\newblock \href {https://proceedings.mlr.press/v32/le14.html} {Distributed
  representations of sentences and documents}.
\newblock In \emph{Proceedings of the 31st International Conference on Machine
  Learning}, volume~32 of \emph{Proceedings of Machine Learning Research},
  pages 1188--1196, Bejing, China. PMLR.

\bibitem[{Mikolov et~al.(2013)Mikolov, Chen, Corrado, and
  Dean}]{mikolov_efficient_2013}
Tomas Mikolov, Kai Chen, G.~Corrado, and J.~Dean. 2013.
\newblock \href {http://arxiv.org/abs/1301.3781} {Efficient {Estimation} of
  {Word} {Representations} in {Vector} {Space}}.
\newblock In \emph{Proceedings of 1st {International} {Conference} on
  {Learning} {Representations}, \{{ICLR}\} 2013}.

\bibitem[{Montanelli and Periti(2023)}]{montanelli2023survey}
Stefano Montanelli and Francesco Periti. 2023.
\newblock \href {http://arxiv.org/abs/2304.01666} {A survey on contextualised
  semantic shift detection}.

\bibitem[{Rehurek and Sojka(2011)}]{rehurek2011gensim}
Radim Rehurek and Petr Sojka. 2011.
\newblock Gensim--python framework for vector space modelling.
\newblock \emph{NLP Centre, Faculty of Informatics, Masaryk University, Brno,
  Czech Republic}, 3(2).

\bibitem[{Sch{\"o}nemann(1966)}]{Schonemann:1966:GSO}
Peter~H. Sch{\"o}nemann. 1966.
\newblock A generalized solution of the orthogonal {Procrustes} problem.
\newblock \emph{j-PSYCHO}, 31(1):1--10.

\bibitem[{Szymanski(2017)}]{szymanski-2017-temporal}
Terrence Szymanski. 2017.
\newblock \href {https://doi.org/10.18653/v1/P17-2071} {Temporal word
  analogies: Identifying lexical replacement with diachronic word embeddings}.
\newblock In \emph{Proceedings of the 55th Annual Meeting of the Association
  for Computational Linguistics (Volume 2: Short Papers)}, pages 448--453,
  Vancouver, Canada. Association for Computational Linguistics.

\bibitem[{Tahmasebi et~al.(2019)Tahmasebi, Borin, and
  Jatowt}]{tahmasebi2019survey}
Nina Tahmasebi, Lars Borin, and Adam Jatowt. 2019.
\newblock \href {http://arxiv.org/abs/1811.06278} {Survey of computational
  approaches to lexical semantic change}.

\end{thebibliography}

\bibliographystyle{acl_natbib}

\appendix

\section{Word analogues through time}
\label{sec:appendixWordAnalogues}

In this appendix, we present the two most similar analogues for each decade related to the studied words. Table \ref{tab:energy_onomasy} focuses on energy-related concepts, Table \ref{tab:transport_onomasy} examines changes in transport concepts, Table \ref{tab:entertainment_onomasy} illustrates analogies related to entertainment, and Table \ref{tab:technology_onomasy} displays analogies linked to computing.

\begin{table*}[htbp]
\centering
\begin{tabular}{lllll}
\toprule
 & petroleum & diesel & electricity & nuclear \\
\midrule
\textbf{1800} & ores, imported & tar, steam & electricity, coals & repelling, blockade \\
\textbf{1810} & merchandise, imported & steam, staves & steam, electricity & war, alliances \\
\textbf{1820} & commodities, imported & 212, tons & electricity, furnaces & chemical, piratical \\
\textbf{1830} & railways, coal & steam, propelled & electricity, electrical & war, eventual \\
\textbf{1840} & shipment, spices & steam, engines & steam, electricity & waging, war \\
\textbf{1850} & guano, ores & boilers, steam & railways, electric & dynamic, explosion \\
\textbf{1860} & metals, mines & boilers, steam & fuel, electricity & forts, destructive \\
\textbf{1870} & coal, ores & steam, engines & fuel, electric & chemical, disintegration \\
\textbf{1880} & coal, mines & boilers, steam & electric, electricity & chemical, projectiles \\
\textbf{1890} & mines, coal & gasoline, turbines & electricity, supply & arsenal, chemical \\
\textbf{1900} & coal, petroleum & engines, generators & electricity, fuel & wireless, resists \\
\textbf{1910} & petroleum, dyers & motors, engines & electricity, electric & aircraft, explosion \\
\textbf{1920} & petroleum, coal & petrol, steam & electricity, fuel & batteries, submarine \\
\textbf{1930} & petroleum, coal & gasoline, boilers & electricity, electric & electronic, submarine \\
\textbf{1940} & petroleum, coal & gasoline, turbine & electricity, electric & nuclear, atomic \\
\textbf{1950} & petroleum, oil & gasoline, diesel & electricity, electric & nuclear, atomic \\
\textbf{1960} & petroleum, api & compressors, diesel & electricity, fuel & nuclear, stockpiles \\
\textbf{1970} & \makecell[l]{petroleum,\\petrochemicals} & diesel, gasoline & electricity, electric & nuclear, weapons \\
\textbf{1980} & petroleum, oil & diesel, gasoline & electricity, electric & nuclear, weapons \\
\textbf{1990} & petroleum, refiners & diesel, gasoline & electricity, electric & nuclear, weapons \\
\bottomrule
\end{tabular}

\caption{\label{tab:energy_onomasy}Temporal word analogies for energy related concepts.
}
\end{table*}

\begin{table*}[htbp]
\centering
\begin{tabular}{lllll}
\toprule
 & bus & truck & train & aircraft \\
\midrule
\textbf{1800} & carriages, coach & cart, jumped & caravan, passengers & ships, privateers \\
\textbf{1810} & lodgings, carriages & waggon, cart & ahead, reconnoitre & ships, bomb \\
\textbf{1820} & waggon, ball & alongside, cart & abreast, ride & ships, ship \\
\textbf{1830} & carriage, passengers & towed, carriage & carriage, boat & ships, vessel \\
\textbf{1840} & gig, coach & gig, towed & coach, ride & ships, engines \\
\textbf{1850} & car, barge & gangway, gig & barge, shoved & ships, ship \\
\textbf{1860} & carriage, car & wagon, carts & train, thither & ships, ship \\
\textbf{1870} & cars, carriage & driver, cart & cab, train & ships, engines \\
\textbf{1880} & cars, carriage & hitched, wagon & cars, train & ship, ships \\
\textbf{1890} & cars, cab & cart, wagon & coach, cab & engines, steamships \\
\textbf{1900} & cars, cab & cart, gigs & train, trains & ships, ship \\
\textbf{1910} & cars, car & cart, wagon & train, car & ships, ship \\
\textbf{1920} & car, train & car, truck & train, trains & aircraft, ships \\
\textbf{1930} & car, bus & car, truck & train, trains & aircraft, ships \\
\textbf{1940} & bus, car & truck, car & train, trains & aircraft, ships \\
\textbf{1950} & bus, car & truck, car & train, trains & aircraft, ships \\
\textbf{1960} & bus, car & truck, car & train, trains & aircraft, takeoff \\
\textbf{1970} & bus, train & truck, car & train, trains & aircraft, ship \\
\textbf{1980} & bus, train & truck, car & train, trains & aircraft, ships \\
\textbf{1990} & bus, buses & truck, car & train, bus & aircraft, ships \\
\bottomrule
\end{tabular}

\caption{\label{tab:transport_onomasy}Temporal word analogies for transport related concepts.
}
\end{table*}

\begin{table*}[htbp]
\centering
\begin{tabular}{llll}
\toprule
 & radio & cinema & television \\
\midrule
\textbf{1800} & spies, circulated & theatres, comic & theatres, exhibited \\
\textbf{1810} & ringing, circulated & beau, stage & newspapers, coaches \\
\textbf{1820} & newspapers, drums & operas, italian & newspapers, operas \\
\textbf{1830} & newspapers, canals & operas, museums & theatres, watches \\
\textbf{1840} & instruments, newspapers & novels, sentimental & concerts, plays \\
\textbf{1850} & harpers, musical & celebrities, academy & actresses, newspapers \\
\textbf{1860} & telegraph, audiences & thiers, museums & locomotive, newspapers \\
\textbf{1870} & electric, telegraph & renaissance, crowe & newspapers, locomotive \\
\textbf{1880} & telephone, telegraph & provencal, forestry & newspapers, reports \\
\textbf{1890} & telephone, newspapers & theater, graeco & newspapers, newspaper \\
\textbf{1900} & telephone, telegraph & studio, theatres & newspapers, theatres \\
\textbf{1910} & telephone, telephones & theatre, theater & newspapers, pictures \\
\textbf{1920} & radio, telephone & theater, theatre & radio, newspapers \\
\textbf{1930} & radio, telephone & cinema, theater & radio, movie \\
\textbf{1940} & radio, wireless & theater, theatres & radio, movies \\
\textbf{1950} & radio, television & romantic, theatre & television, radio \\
\textbf{1960} & radio, television & theater, comics & television, tv \\
\textbf{1970} & radio, television & cinema, theatre & television, tv \\
\textbf{1980} & radio, television & cinema, movie & television, tv \\
\textbf{1990} & radio, television & cinema, feminism & television, tv \\
\bottomrule
\end{tabular}

\caption{\label{tab:entertainment_onomasy}Temporal word analogies for entertainment related concepts.
}
\end{table*}

\begin{table*}[htbp]
\centering
\begin{tabular}{llll}
\toprule
 & computer & internet & email \\
\midrule
\textbf{1800} & mathematical, perspective & information, access & gazette, messages \\
\textbf{1810} & mathematical, surgery & retail, information & courier, newspapers \\
\textbf{1820} & sciences, laboratory & information, enables & advertise, requesting \\
\textbf{1830} & science, manual & information, access & letter, sent \\
\textbf{1840} & sciences, mathematical & available, accessible & forwarded, courier \\
\textbf{1850} & orally, mathematical & information, access & forwarded, letter \\
\textbf{1860} & operator, experimental & accessible, telegraph & courier, enquirer \\
\textbf{1870} & telegraphy, engineering & facilities, valuable & courier, forwarded \\
\textbf{1880} & chemical, engine & access, information & lippincott, forwarded \\
\textbf{1890} & textbooks, laboratory & accessible, information & forwarded, blackwood \\
\textbf{1900} & electro, arithmetic & information, accessible & sent, send \\
\textbf{1910} & automobile, classroom & mail, information & forwarded, prepaid \\
\textbf{1920} & classroom, equipment & opportunities, information & forwarded, telegrams \\
\textbf{1930} & electrical, dynamo & information, access & forwarded, phone \\
\textbf{1940} & blackboard, computer & access, telephone & notices, send \\
\textbf{1950} & computer, digital & television, information & addresses, telegrams \\
\textbf{1960} & computer, computers & users, information & telegrams, penguin \\
\textbf{1970} & computer, computers & information, computers & mailing, telegram \\
\textbf{1980} & computer, computers & computer, computers & mail, telex \\
\textbf{1990} & computer, computers & internet, web & email, fax \\
\bottomrule
\end{tabular}

\caption{\label{tab:technology_onomasy}Temporal word analogies for computing related concepts.
}
\end{table*}

\end{document}